\documentclass{article}
\usepackage[a4paper,margin=2cm]{geometry}
\usepackage[colorlinks]{hyperref}

\usepackage[
activate={true,nocompatibility}, 
tracking=true,
kerning=true,
spacing=true,
stretch=10,
shrink=10]{microtype} 

\usepackage{amsmath}
\usepackage{amssymb}
\usepackage{caption}

\usepackage{multirow}

\usepackage{tikz}
\usepackage{pgfplots}
\pgfplotsset{compat=newest}
\usepgfplotslibrary{groupplots}

\usepackage{authblk}

\newlength{\colw}
\usepackage{xspace}
\newcommand{\modelname}{MeFEm\xspace}
\newcommand{\NA}{---}

\author[1]{Yury Borets\thanks{boretsyury@gmail.com}}
\author[1]{Stepan Botman\thanks{stepan.botman@gmail.com}}
\affil[1]{Sber AI Lab}

\begin{document}
\title{MeFEm: Medical Face Embedding model}

\maketitle

\begin{abstract}
We present \modelname, a vision model based on a modified Joint Embedding Predictive Architecture (JEPA) for biometric and medical analysis from facial images.
Key modifications include an axial stripe masking strategy to focus learning on semantically relevant regions, a circular loss weighting scheme, and the probabilistic reassignment of the CLS token for high quality linear probing.
Trained on a consolidated dataset of curated images, \modelname outperforms strong baselines like FaRL and Franca on core anthropometric tasks despite using significantly less data.
It also shows promising results on Body Mass Index (BMI) estimation, evaluated on a novel, consolidated closed-source dataset that addresses the domain bias prevalent in existing data.
Model weights are available at \url{https://huggingface.co/boretsyury/MeFEm}, offering a strong baseline for future work in this domain.
\end{abstract}

\section{Introduction}

The rise of large-scale foundational models has demonstrated unprecedented capabilities across language~\cite{naveed2025comprehensive} and vision~\cite{awais2025foundation} tasks, typically trained on broad internet-scale data.
A compelling, yet underexplored, application of this paradigm lies in medicine~\cite{khan2025comprehensive}, where significant progress has been made despite a severe constraint: the scarcity of suitable training data.
The utility of a foundational model depends fundamentally on the volume of high-quality data, and the poor generalization of general-purpose models on specialized medical tasks is a well-known challenge (e.g., \cite{ma2024segment}).

Facial images, in contrast, represent a uniquely powerful and accessible modality. They are ubiquitous, easily captured, and rich in physiological information that can reflect systemic health status and potential underlying pathologies.
Datasets of facial images annotated with medical labels are typically small, proprietary, or both.
This bottleneck arises from two fundamental challenges: the stringent privacy and ethical concerns surrounding the sharing of sensitive biometric data like faces, and the significant cost and expertise required for accurate data labeling, which often involves laboratory tests, specialized instrumentation, or access to confidential medical records.

This work addresses this challenge by proposing a bridge between the two domains: we leverage self-supervised learning (SSL) on large-scale, non-medical facial image collections to build a model that captures robust, medically-relevant features.
The core hypothesis is that a model trained to understand the fundamental structure and variation of human faces through SSL will learn representations that are highly effective for downstream medical and biometric prediction tasks, even with limited labeled data.
This approach bypasses the need for direct, sensitive medical labels during pretraining while still aiming to capture the subtle phenotypic markers associated with health states.

\section{Related works}
\subsection{Self-supervised learning}

SSL provides a powerful framework for learning representations from unlabeled data by defining a pretext task where the target is generated from the data itself.
Early context-based methods, such as predicting image rotations or solving jigsaw puzzles, demonstrated promise but often learned features that did not transfer well to high-level semantic tasks.
The field was subsequently dominated by contrastive learning paradigms (e.g., SimCLR~\cite{chen2020simclr}, MoCo~\cite{he2020mocov1, chen2021mocov3}), which learn by maximizing agreement between differently augmented views of the same image while pushing apart views of different images, and self-distillation methods (e.g., BYOL~\cite{grill2020byol}, DINO~\cite{caron2021dino, oquab2023dinov2}), which avoid negative samples by having a student network predict the output of a teacher network derived from its own past states.
While highly effective, these methods can be computationally expensive and require careful management of negative samples or sophisticated asymmetric architectures to avoid collapse.

A more recent and influential shift has been towards generative masked modeling, where a model learns by predicting masked portions of the input.
These methods are primarily distinguished by their prediction target.
Some, like Masked Autoencoders~\cite{he2022mae} and SimMIM~\cite{xie2022simmim}, perform pixel-level reconstruction.
In contrast, our work builds upon methods that perform representation-level prediction, aiming to predict the latent features of masked regions rather than raw pixels. 
This category includes frameworks like data2vec~\cite{baevski2022data2vec} and the Joint Embedding Predictive Architecture (JEPA)~\cite{assran2023self}.
These representation-based methods inherit a core principle from contrastive and self-distillation learning—the importance of learning in a structured latent space—but implement it through a predictive, masked-modeling objective. 
This approach is more computationally efficient and encourages the model to capture higher-level semantics. 
The JEPA framework, in particular, has proven to be a highly scalable and versatile architecture, forming the foundation for state-of-the-art models in video understanding~\cite{assran2025vjepa2} and other modalities. Its ability to learn robust representations by predicting in latent space makes it a premier choice for domains where semantic understanding is paramount over pixel-level fidelity.


\subsection{Embedding models}

The landscape of computer vision has been reshaped by the release of powerful open-weight foundational models.
These models, often Vision Transformer~(ViT)~\cite{dosovitskiy2020image} based, are characterized by their scale, training data, and—critically—their public availability, serving as standardized feature extractors and benchmarks.
This ecosystem is divided into distinct paradigms based on their learning objective.

One dominant paradigm relies on language supervision, using image-text pairs.
This includes contrastive models like Open-CLIP~\cite{ilharco2021openclip} and generative Vision-Language Models (VLMs) like CoCa~\cite{yu2022coca} and Emu~\cite{sun2023emu}.
While revolutionary for general-purpose reasoning, their representational space is optimized for semantic alignment with text, which may not capture the fine-grained, sub-semantic features crucial for specialized domains.

In parallel, a suite of purely visual foundational models has been established through SSL.
This includes models like DINOv2 and I-JEPA, which learn powerful representations from images alone, without any text-based guidance.
A notable recent advancement in this line is the Franca model~\cite{venkataramanan2025franca}, which scales this SSL approach to billions of parameters, establishing a new state-of-the-art for general visual recognition.

The success of these foundational encoders has inspired their adaptation to specialized domains.
A key example is FaRL~\cite{zheng2022general}, which applies a CLIP-like, image-text contrastive objective to a massive dataset of faces to learn a general-purpose facial encoder.
Our work builds upon this progression but pursues an alternative path.
We hypothesize that for specialized facial analysis tasks where textual alignment offers no inherent advantage, further specializing a purely visual SSL approach can yield superior performance by focusing the model's capacity on fine-grained visual features rather than text-aligned semantics.

\section{Materials and methods}

This work employs the standard ViT architecture as its encoder backbone, consistent with the original JEPA implementation and other foundational models.
While subsequent hierarchical architectures have been developed to address the fixed-scale receptive field of the vanilla ViT, we demonstrate that a targeted preprocessing strategy effectively resolves this issue for structured facial data.
By standardizing input images through rigorous scaling and cropping, we enforce a consistent spatial alignment between facial features and the model's patch grid.
This ensures the fixed receptive field becomes a predictable and stable basis for feature extraction rather than a limitation, thereby justifying the use of ViT architecture for our domain.

We treat face detection as a solved preliminary step, utilizing established models to form the foundation of our preprocessing pipeline.
In this first stage, we employ the Blazeface model~\cite{bazarevsky2019blazeface} from MediaPipe to obtain a bounding box, which is then used to crop image, ensuring the face is centered and scaled to consistent proportions.
This directly addresses the ViT's fixed-scale constraint by creating a stable correspondence between anatomical landmarks and the input patches.

Standard face detection models typically produce a tight bounding box, spanning from chin to brow and cheek to cheek.
While computationally efficient, this approach excludes potentially informative regions such as the forehead, ears, and neck.
To retain this contextual information, we define our final crop as a square with dimensions twice the maximum dimension (height or width) of the original bounding box, centered on the detected face.
This strategy ensures the inclusion of peripheral anatomical features without introducing distortion.
In practice, this method produces stable and consistent inputs, with only minor shifts due to variations in camera angle or head pose.


\subsection{Data}
\subsubsection{Training dataset}

A primary source of open-access facial data is the FaceCaption-15M dataset~\cite{dai2024facecaption15m}, which provides images alongside textual descriptions (unused in utilized self-supervised approach) and face bounding boxes, which we utilized directly, bypassing the need to run our own detection model.
This dataset is distributed as a list of hyperlinks and, as its authors noted regarding the parent LAION-Faces-50M, suffers from inherent link rot problem.
From the accessible images, we performed a cleaning procedure that removed samples where a face could not be properly cropped (e.g. due to proximity to the image boundary) or where the resulting crop resolution was below \(224\times224\) pixels.
The final subset amounts to approximately one-third of the original, totaling under 4.6~M samples, and forms the backbone of our training set, as detailed in Table~\ref{tab:datasets_train}.

To enrich our training data, we incorporated samples from the AVSpeech dataset~\cite{ephrat2018looking}, which consists of short video clips of people speaking.
These clips are also distributed via hyperlinks and are consequently subject to the same link rot issue.
Since each clip features a single speaker, we extracted one frame per clip to maximize data diversity and avoid redundancy.
We processed the resulting images using the same pipeline applied to FaceCaption-15M, yielding an additional 1.5~M samples that provide more 'in-the-wild' variation.

\begin{figure}[htb]
\newcommand{\imageblockspacing}{5pt}
\centering
\setlength{\tabcolsep}{\imageblockspacing}
\begin{tabular}{c|ccc}
& SHFQ & AVSpeech & FaceCaption \\
\hline
Good samples &
\begin{minipage}{0.245\textwidth}
\vspace{\imageblockspacing}
\centering
\setlength{\tabcolsep}{0pt}
\renewcommand{\arraystretch}{0}
\begin{tabular}{cccc}
\includegraphics[width=0.25\linewidth]{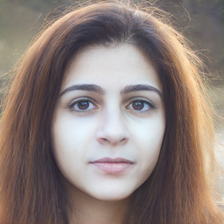} &
\includegraphics[width=0.25\linewidth]{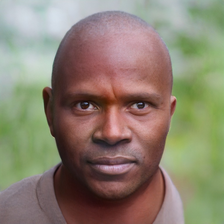} &
\includegraphics[width=0.25\linewidth]{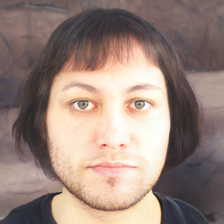} &
\includegraphics[width=0.25\linewidth]{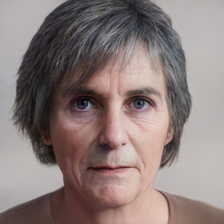} \\
\end{tabular}
\vspace{\imageblockspacing}
\end{minipage} &
\begin{minipage}{0.245\textwidth}
\vspace{\imageblockspacing}
\centering
\setlength{\tabcolsep}{0pt}
\renewcommand{\arraystretch}{0}
\begin{tabular}{cccc}
\includegraphics[width=0.25\linewidth]{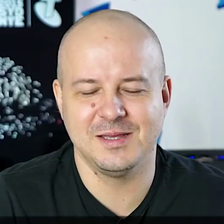} &
\includegraphics[width=0.25\linewidth]{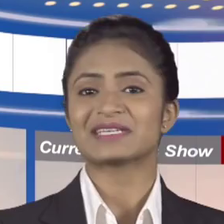} &
\includegraphics[width=0.25\linewidth]{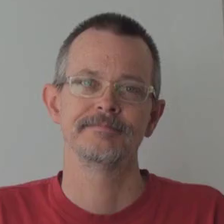} &
\includegraphics[width=0.25\linewidth]{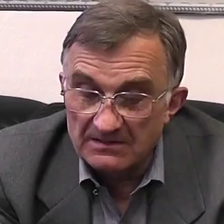} \\
\end{tabular}
\vspace{\imageblockspacing}
\end{minipage} &
\begin{minipage}{0.245\textwidth}
\vspace{\imageblockspacing}
\centering
\setlength{\tabcolsep}{0pt}
\renewcommand{\arraystretch}{0}
\begin{tabular}{cccc}
\includegraphics[width=0.25\linewidth]{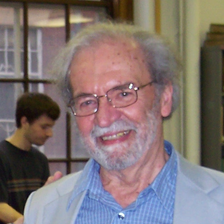} &
\includegraphics[width=0.25\linewidth]{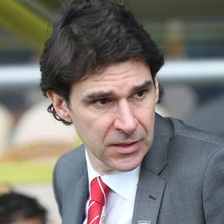} &
\includegraphics[width=0.25\linewidth]{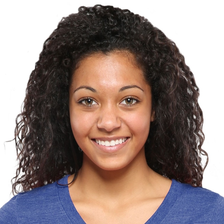} &
\includegraphics[width=0.25\linewidth]{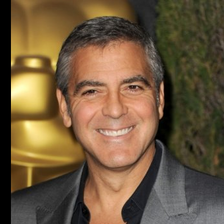} \\
\end{tabular}
\vspace{\imageblockspacing}
\end{minipage} \\
Bad samples & &
\begin{minipage}{0.245\textwidth}
\vspace{\imageblockspacing}
\centering
\setlength{\tabcolsep}{0pt}
\renewcommand{\arraystretch}{0}
\begin{tabular}{cccc}
\includegraphics[width=0.25\linewidth]{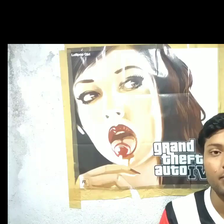} &
\includegraphics[width=0.25\linewidth]{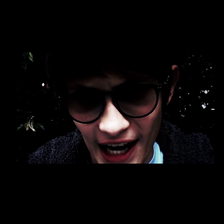} &
\includegraphics[width=0.25\linewidth]{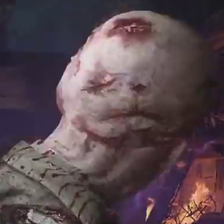} &
\includegraphics[width=0.25\linewidth]{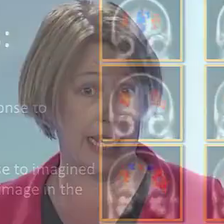} \\
\end{tabular}
\vspace{\imageblockspacing}
\end{minipage} &
\begin{minipage}{0.245\textwidth}
\vspace{\imageblockspacing}
\centering
\setlength{\tabcolsep}{0pt}
\renewcommand{\arraystretch}{0}
\begin{tabular}{cccc}
\includegraphics[width=0.25\linewidth]{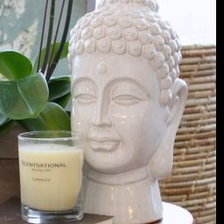} &
\includegraphics[width=0.25\linewidth]{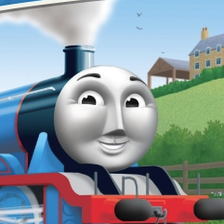} &
\includegraphics[width=0.25\linewidth]{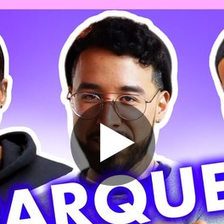} &
\includegraphics[width=0.25\linewidth]{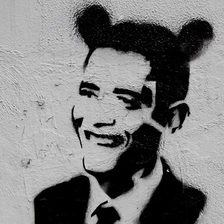} \\
\end{tabular}
\vspace{\imageblockspacing}
\end{minipage} \\
\end{tabular}
\caption{Visual examples from the datasets comprising the training set.}
\label{fig:training_images}
\end{figure}

As illustrated in Figure~\ref{fig:training_images}, both the AVSpeech and FaceCaption datasets contain a minor proportion of irrelevant or mislabeled samples, an inevitable consequence of their scale and automated collection pipelines.
We estimate these constitute a small single-digit percentage of the total data.
Although filtering these samples would likely improve downstream performance, developing a robust detection method is a non-trivial task that lies beyond the scope of this work.

Finally, we incorporated the SFHQ dataset~\cite{beniaguev2022SFHQ}, which contains 0.5~M synthetically generated, high-quality face images at a resolution of \(1024\times1024\) pixels.
This dataset provides high variability in age, ethnicity, and facial expression.
Since the faces are pre-centered and cropped, the only required preprocessing was downscaling the images to our target resolution.

\begin{table}[!htb]
\centering
\begin{tabular}{c|c|c|c}
Dataset name & Original Size & Media type & Usable size  \\
\hline
FaceCaption-15M & 15~078~935 & images &  4~572~807 \\
AVSpeech & \hphantom{0}2~621~845 & videos & 1~446~778 \\
SFHQ & \hphantom{00~}425~258 & images & \hphantom{0~}425~258 \\
\hline 
\multicolumn{3}{r|}{Total} & 6~444~843\\
\end{tabular}
\caption{Composition of training dataset. The 'usable' refers to samples that were both accessible online and met face-cropping and resolution criteria.}
\label{tab:datasets_train}
\end{table}

\subsubsection{Evaluation datasets}

A direct assessment of medical screening proficiency would require datasets of facial images labeled with specific diseases, which are scarce and rarely publicly available.
Consequently, we evaluate a critical prerequisite for such tasks: the accurate prediction of anthropometric and demographic parameters from images.
This approach is justified for two reasons.
First, several public datasets provide established tasks for age, gender, and ethnicity prediction, enabling a direct comparison of model capabilities.
Second, these parameters are themselves well-established biomarkers, strongly correlated with morbidity, mortality, and overall health status.
A model that excels at estimating these features demonstrates a core competency that is directly relevant to downstream clinical applications.

For evaluation, we utilize the CelebA dataset~\cite{liu2015faceattributes}, using its provided attribute labels as prediction targets. For age classification, we employ the FairFace dataset~\cite{karkkainen2021fairface}, structuring the task by predicting gender within separate 10-year age bins for white and non-white subject groups.

A core component of our evaluation is the estimation of Body Mass Index~(BMI), a physiological parameter of primary clinical importance, alongside age, gender, and ethnicity.
Predicting BMI from facial morphology is notoriously difficult, and while state-of-the-art models can achieve a Mean Absolute Error~(MAE) within the 2–-5 range on specific datasets~\cite{siddiqui2022examination}, their performance deteriorates significantly under domain shift.
This fragmentation across numerous small datasets leads to models that fail to generalize.
To address this, we constructed a new, consolidated BMI dataset to enable a more robust and reliable assessment of model performance on this crucial task.
The final consolidated BMI dataset comprises three sources: FIW-BMI~\cite{jiang2019visual} (7769 images), MCD-rPPG~\cite{egorov2025gaze} (750 images), and a custom-collected dataset (4784 images).
The custom dataset was assembled by curating publicly available images with user-reported height and weight. 
While this source provides valuable diversity in subject demographics, body types, and imaging conditions, the data cannot be shared publicly due to its nature.
Finally, we leverage the instrumentally and laboratory-measured biomarkers provided in the MCD-rPPG dataset to conduct a preliminary assessment of our model's performance on more challenging, clinically-grounded prediction tasks.


\subsection{Training process}
\subsubsection{JEPA basics}

The core idea of JEPA approach is predicting representation of some image portion in latent space based on information in the remaining parts of image.
That is achieved by splitting set of image patch tokens \(\mathcal{P} = \{p_1, p_2, \ldots, p_N\}\) into two subsets: source \(\mathcal{P}_S\) and target \(\mathcal{P}_T\) so that \(\mathcal{P}_S \cup \mathcal{P}_T = \mathcal{P}\) and \(\mathcal{P}_S \cap \mathcal{P}_T = \emptyset\).
While split can done in purely random fashion, in JEPA split is structured so that both subsets contain relatively large continuous segments of original image which allows model to learn not only local features but also long-range dependencies.
The source tokens $\mathcal{S}$ and target tokens $\mathcal{T}$ are passed through separate encoders \(\text{Enc}_{\scriptscriptstyle S}\) and \(\text{Enc}_{\scriptscriptstyle T}\), which share same architecture.
Final piece of this approach is predictor \(\text{Pred}\) which reconstruct latent representation of targets tokens from source ones, which can be compared to what was obtained from target encoder:  
\begin{equation}
\mathcal{L} = \frac{1}{|\mathcal{P}_T|} \sum \limits_{p_i \in\mathcal{P}_T} \left\Vert \vphantom{\int} \text{Pred} \left( \text{Enc}_{\scriptscriptstyle S}(\mathcal{P}_S) \right) - \text{Enc}_{\scriptscriptstyle T}(\mathcal{P}_T) \right\Vert_2^{i},
\label{eq:basic_jepa_loss}
\end{equation}
where \(\mathcal{L}\) is \(L_2\) norm based loss function used in training loop (technically smooth \(L1\) is used in practice instead, but this distinction is not essential for the following reasoning).
Weights of source encoder and predictor are updated the usual way using error backpropagation, while target encoder weights are updated as exponentially averaged source encoder weights.
This stabilizes training dynamics and impede representation collapse.

\subsubsection{Probabilistic CLS token assignment}  

The original implementation utilized vanilla ViT for patch token encoder, with the only notable difference that CLS (class) token being disabled.
Omitting CLS token certainly streamlines architecture and training process and while the reason was not explicitly stated in JEPA paper it's a natural choice for fundamental model. 
While the CLS token is commonly used in transformers for image data and beyond with new methods regularly proposed~\cite{liu2023simple, li2024multi} to enrich its expressiveness, we suggest that JEPA training setup provide sufficient pressure for CLS token to learn meaningful representation with only minimal modifications to it. 

Let \(c\) be CLS token, with complete token set for image denoted as \(\mathcal{X} = \mathcal{P} \cup \{c\}\).
Since masking strategy operates in term of patch positions --- a property CLS token lacks --- we propose to append it to either source or target sets in probabilistic manner, independently of image patches tokens:
\begin{equation}
\mathcal{S} = 
\begin{cases}
\mathcal{P}_S \cup \{c\} & \text{if } b = 1 \\
\mathcal{P}_S & \text{if } b = 0
\end{cases}, \quad
\mathcal{T} = 
\begin{cases}
\mathcal{P}_T & \text{if } b = 1 \\
\mathcal{P}_T \cup \{c\} & \text{if } b = 0
\end{cases},
\end{equation}
where \(b\) is obtained by sampling from Bernoulli distribution \(b \sim \mathrm{Bernoulli}(P_{c \in \mathcal{S}})\)
where \(P_{c \in \mathcal{S}} \in [0, 1]\) --- probability of assigning CLS token to source set.
The loss is then calculated as usual: 
\begin{equation}
\mathcal{L} = \frac{1}{|\mathcal{T}|} \sum \limits_{t_i \in \mathcal{T}} \left\Vert \vphantom{\int} \text{Pred} \left( \text{Enc}_{\scriptscriptstyle S}(\mathcal{S}) \right) - \text{Enc}_{\scriptscriptstyle T}(\mathcal{T}) \right\Vert_2^i
\label{eq:jepa_cls_loss}
\end{equation}

For most of our experiments we use \(P_{c \in \mathcal{S}} = 0.5\) as a balanced choice, which enables the CLS token to learn a meaningful global representation without compromising the detailed, localized learning of the patch embeddings.


\subsubsection{Axial stripe masking}  
One of the core alterations to the original JEPA approach we introduce stems from the fact that we operate on images structured in a certain way, particularly a face centered within frame boundaries.
This leads to the necessity of introducing domain-specific masking strategy, namely axial stripe masking.

\begin{figure}[htb]
\centering
\begin{tikzpicture}
\begin{groupplot}[
    scale=0.65,
    group style={
        group size=3 by 1,
        horizontal sep=1cm,
        vertical sep=1cm
    },
    axis equal image,
    axis lines = middle,
    xmin=0, xmax=14,
    ymin=0, ymax=14,
    xtick={0,1,...,15},
    ytick={0,1,...,15},
    xticklabels=\empty,
    yticklabels=\empty,
    tick style={draw=none},
    grid=major,
    major grid style={line width=0.5pt, draw=gray!50},
    axis line style={line width=0.5pt, draw=gray!50, -},
]
\nextgroupplot
\fill[yellow!50, opacity=0.5, even odd rule]
    (0,0) rectangle (15,15)
    (1,1) -- (9,1) -- (9, 5) -- (12, 5) -- (12, 12) -- (5, 12) -- (5, 9) -- (1,9) -- cycle;  
\nextgroupplot
\fill[green!30, opacity=0.75] (0,0) rectangle (7,7);
\nextgroupplot
\fill[red!30, opacity=0.5] (0,2) rectangle (15,5);
\fill[blue!30, opacity=0.5] (7,0) rectangle (10,15);
\end{groupplot}
\node[anchor=north] at (group c1r1.south) {(a)};
\node[anchor=north] at (group c2r1.south) {(b)};
\node[anchor=north] at (group c3r1.south) {(c)};
\end{tikzpicture}
\caption{ Schematic representation of different masking strategies: multiblock~(a), quadrant~(b) and, axial stripes~(c). Colored regions represent source the mask; remaining part is the target.}
\label{fig:masking_strategies}
\end{figure}
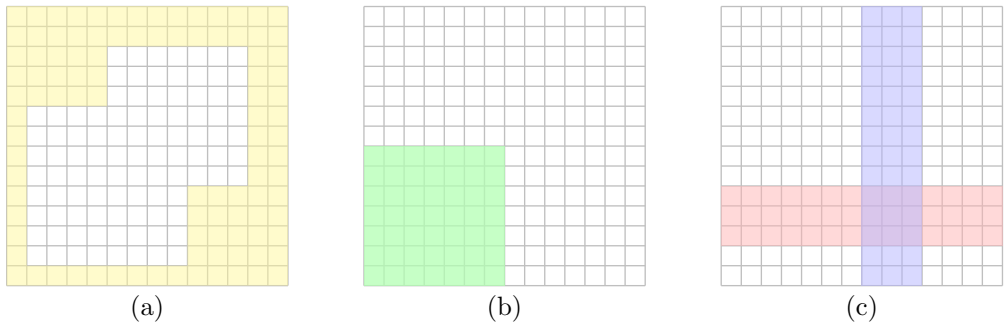

In multiblock masking strategy utilized in JEPA, the final target mask is formed by merging set of rectangular masks with sizes, aspect ration and positions sampled from predetermined intervals (see Figure~\ref{fig:masking_strategies}~a).
Particularly, two modes are used simultaneously during training: one with small number of large masks and another with large number of small masks.
A direct consequence of this strategy is that image patches have non-uniform probability of being assigned to the target subset which becomes function of their spatial positions.
Patches near image edges are more likely to be included into source mask as they are less likely to be covered by a randomly placed mask.
This is self evident for mask larger than half image in size but remains true even for the case of smaller masks.
This positional bias creates a fundamental problem for our domain as the most semantically relevant image portions are concentrated in the center of the image, while background portions that are irrelevant for our purposes are located near edges. 
While the original JEPA avoids such issues through the use of random crop augmentations, we strive to preserve the consistent spatial structure of face images.
This necessitates an alternative, more appropriate masking solution.

The core requirement for the new masking strategy is that the source mask must, by design, contain patches relevant for reconstructing facial representation, with background patches being optional.
While quadrant masking can easily fulfill this requirement (see Figure~\ref{fig:masking_strategies}~b), it lacks flexibility to control source-target mask ratio during training.
This negatively impacts metrics, rendering it suboptimal choice for the problem at hand.

Thus, we introduce a new masking strategy --- axial stripe masking --- which defines source mask as either horizontal or vertical stripe spanning full image (see Figure~\ref{fig:masking_strategies}~c).
In contrast to a quadrant mask, its area can be controlled via width parameter and, unlike multiblock masking, these full-height or full-width are designed to consistently intersect semantically relevant regions, ensuring source context always contain meaningful information.

We enhance this approach even further by sampling stripe positions~\(i_c\) from Gaussian distribution centered on the image midpoint:
\begin{equation}
i_c = \text{round} \left( \mathcal{N}_{[0, L-1]}\left( \frac{L - 1}{2}, L^2 \, k^2 \right) \right)
\end{equation}
where \(\mathcal{N}_{[a, b]} \left( \mu, \sigma^2 \right)\) --- is truncated normal distribution, \(L=14\) is number of patches per axis, and coefficient \( k=0.175\) was chosen empirically.

During training, the stripe's direction and position are chosen randomly for each sample, while width is fixed at 3 patches, a value found to be optimal.
When CLS token is appended to the source set, the last image patch is dropped from the mask to maintain  consistent shape across the batch, which provides a slight training speedup.
Furthermore, the dropped patch is intentionally located at the image border making it safe to remove. 

\subsubsection{Circular loss weighs}  

The final improvement we propose aims to reduce the contribution of irrelevant patches to the total loss.
While we lack per-patch annotations for our unlabeled data, and full image segmentation, while possible, would be computationally expensive, we argue for a more efficient alternative.

The argument is that significant advantage can be achieved by down-weighting the loss from irrelevant patches, even via a crude approximation.
We propose to do this by weighting each patch's loss based on its distance from image center:  
\begin{equation}
\mathcal{L} = \frac{1}{|\mathcal{T}|} \sum \limits_{t_i \in \mathcal{T}} w_i \cdot \left\Vert \vphantom{\int} \text{Pred} \left( \text{Enc}_{\scriptscriptstyle S}(\mathcal{S}) \right) - \text{Enc}_{\scriptscriptstyle T}(\mathcal{T}) \right\Vert_2^i
\label{eq:jepa_weighted_loss}
\end{equation}
where \(w_i\) --- element of weight matrix and \(w_i = 1\) for \(t_i \equiv c\).
For the weight matrix construction, we selected sigmoid function because it provides a flat region for center patches and smooth falloff towards the edges:
\begin{equation}
w(r) = \frac{1}{1 + \exp \left( \sigma (r_0 - r) \right)},
\end{equation}
where we implied that \(w_i = w(x, y) = w(r)\) and \(r = \left\Vert x - x_0, y - y_0 \right\Vert_2 \) is distance from image center \((x_0, y_0)\).


\begin{figure}[htb]
\centering
\begin{tikzpicture}
\begin{axis}[
    view={0}{90},
    axis equal image,
    width=8.5cm,
    axis line style={draw=none},
    xmin=0.5, xmax=14.5,
    ymin=0.5, ymax=14.5,
    ticks=none,
    point meta min=0,
    point meta max=1,
    colormap/viridis,
    ]
\addplot3[
    matrix plot,
    mesh/cols=14,
    mesh/rows=14,
] table[header=true] {weights.dat};
\addplot3[
    scatter,
    only marks,
    mark=none,
    nodes near coords,
    nodes near coords style={
        font=\tiny,
        color=white,
        text opacity=1,
        anchor=center,
    },
    point meta=explicit symbolic,
] table[meta=label, header=true] {weights.dat};
\end{axis}
\end{tikzpicture}
\caption{Loss weights matrix.}
\label{fig:patch_loss_weights}
\end{figure}
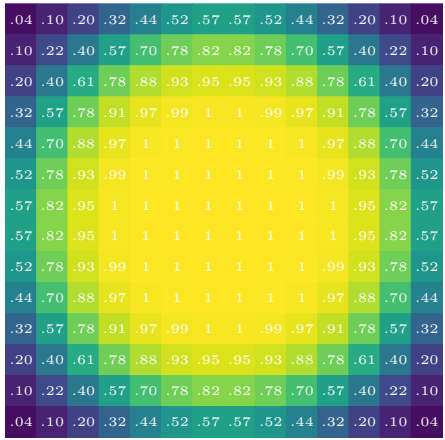




\subsection{Evaluation process}
In order to evaluate the advantages of our proposed model~\modelname, we compare it against three strong baselines: FaRL~(ViT base) and Franca~(ViT large) and Buffalo~(ResNet50).
FaRL is the natural choice as it is the only existing foundational model specifically designed and trained for target domain.
Franca represents the state-of-the-art general-purpose image foundational models; to compensate for inherent domain shift, we use its large variant instead of the base one.
Buffalo refers to the encoder from the Buffalo-L model pack from InsightFace and represents a classic baseline.
It is based on a ResNet50 architecture and has been trained using ArcFace~\cite{deng2018arcface} on the WebFace600K dataset. 
In terms of parameter count, it is closer to the small variants of ViT. All ViT-based models, including our own, use an identical input resolution, which further simplifies the comparison.

To evaluate the quality of representation on downstream tasks, we employ two probing methods: an attentive pooler (following the JEPA implementation) for per-patch embeddings, and a perceptron with a single hidden layer for CLS token only input case.

\section{Results}
The results for all models on the three downstream evaluation tasks are presented in Table~\ref{tab:results_core}.
\modelname model demonstrates a consistent advantage over the baselines across all tasks.
Notably, the performance of CLS token based probing is only a few percentage points lower that the full attentive pooler, while significantly outperforming baseline models.
In this table, as in all subsequent ones, both  the 'Patches' and 'Patches + CLS' configurations use the same attentive poller.
The sole difference is whether CLS token is included in the input.

\begin{table}[htb]
\resizebox{\textwidth}{!}{
\centering
\setlength{\colw}{1.4cm}
\begin{tabular}{c|c|c|c|c|c|c|c}
\hline
\multirow{2}{*}[-6pt]{Model} & \multirow{2}{*}[-6pt]{Size} & \multirow{2}{*}[-6pt]{Features} & \multicolumn{2}{c|}{BMI dataset} & CelebA & \multicolumn{2}{c}{FairFace (age)} \\
\cline{4-8}
& & & \makebox[\colw][c]{\(r^2\)} & \makebox[\colw][c]{MAE} & \makebox[\colw][c]{Acc.} & \makebox[\colw][c]{Acc. (W)} & Acc. (NW) \\
\hline
Franca & Large & Patches + CLS & 0.425 & 3.744 & 0.904 & 0.538 & 0.529 \\
FaRL & Base & Patches & 0.380 & 3.924 & 0.891 & 0.488 & 0.478 \\
FaRL & Base & CLS & 0.445 & 3.604 & 0.904 & 0.563 & \textbf{0.574} \\
Buffalo & Small\(^*\) & CLS & 0.477 & 3.578 & 0.888 & 0.498 & 0.522 \\
\modelname & Base & Patches & \textbf{0.506} & \textbf{3.453} & \textbf{0.910} & \textbf{0.572} & \textbf{0.575} \\
\modelname & Base & CLS & 0.488 & 3.479 & 0.907 & 0.550 & 0.560 \\
\modelname & Base & Patches + CLS & 0.496 & 3.495 & \textbf{0.910} & \textbf{0.570} & \textbf{0.576} \\
\end{tabular}
}
\caption{ Metrics for core downstream tasks compared to baseline models. Note:~\(^*\) indicates the approximate parameter count for a comparable non-ViT model.}
\label{tab:results_core}
\end{table}

In next step we tested \modelname model by predicting essential medical parameters from single frame of MCD-rPPG dataset and compared results against baseline provided in~\cite{egorov2025gaze}.
As expected, the model failed to predict cholesterol levels and diastolic pressure.
For glycated hemoglobin and oxygen saturation, the results indicate some meaningful predictive capability, while metrics for total hemoglobin and systolic pressure warrant cautious optimism.
We interpret these findings as follows: while a single image frame should be insufficient to directly predict parameters acquired via laboratory analysis or specialized measuring instrument, model appears to learn some associations between visual appearance and underlying pathological states, trying to discern subtle changes linked to different types of physiological abnormalities.
The baseline in Table~\ref{tab:results_medical} is based on remote photopletismography (rPPG), which reconstructs the pulse wave from the entire video sequence.
The relatively low metrics for both baseline and \modelname emphasize the inherent difficulty and potential limits of solvability of this problem from visual data alone.

\begin{table}[htb]
\resizebox{\textwidth}{!}{
\centering
\setlength{\colw}{1.0cm}
\begin{tabular}{c|c|c|c|c|c|c|c|c}
\hline
\multirow{2}{*}[-6pt]{Model} & \multirow{2}{*}[-6pt]{Size} & \multirow{2}{*}[-6pt]{Features} & \multicolumn{2}{c|}{\shortstack{Hemoglobin\\(mmol/l)}} & \multicolumn{2}{c|}{\shortstack{\\Glycated\\hemoglobin (\%)}} & \multicolumn{2}{c}{\shortstack{Cholesterol\\(mmol/l)}} \\
\cline{4-9}
& & & \makebox[\colw][c]{\(r^2\)} & \makebox[\colw][c]{MAE} & \makebox[\colw][c]{\(r^2\)} & \makebox[\colw][c]{MAE} & \makebox[\colw][c]{\(r^2\)} & \makebox[\colw][c]{MAE} \\
\hline
MeFEm & Base & Patches & 0.233 & 0.948 & 0.089 & 0.47 & -0.060 & 0.621 \\
MeFEm & Base & CLS & 0.263 & 0.952 & 0.076 & 0.49 & -0.084 & 0.640 \\
MeFEm & Base & Patches + CLS & 0.234 & 0.959 & 0.055 & 0.48 & -0.015 & 0.623 \\
Baseline & \NA & Video superpixels & \NA & \NA & \NA & 0.41 & \NA & 0.60\hphantom{0} \\
\end{tabular}
}

\vspace{0.5cm}

\resizebox{\textwidth}{!}{
\centering
\setlength{\colw}{1.0cm}
\begin{tabular}{c|c|c|c|c|c|c|c|c}
\hline
\multirow{2}{*}[-6pt]{Model} & \multirow{2}{*}[-6pt]{Size} & \multirow{2}{*}[-6pt]{Features} & \multicolumn{2}{c|}{\shortstack{\\Systolic\\pressure (mmHg)}} & \multicolumn{2}{c|}{\shortstack{\\Diastolic\\pressure (mmHg)}} & \multicolumn{2}{c}{\shortstack{\\Saturation (\%)}} \\
\cline{4-9}
& & & \makebox[\colw][c]{\(r^2\)} & \makebox[\colw][c]{MAE} & \makebox[\colw][c]{\(r^2\)} & \makebox[\colw][c]{MAE} & \makebox[\colw][c]{\(r^2\)} & \makebox[\colw][c]{MAE} \\
\hline
MeFEm & Base & Patches & 0.136 & 10.8 & -0.060 & 7.15 & \hphantom{-}0.073 & 0.8 \\
MeFEm & Base & CLS & 0.058 & 11.8 & -0.090 & 7.20 & -0.557 & 1.3 \\
MeFEm & Base & Patches + CLS & 0.115 & 11.2 & -0.054 & 7.14 & \hphantom{-}0.081 & 0.8 \\
Baseline & \NA & Video superpixels & \NA & 12.8 & \NA & 8.39 & \NA & 0.9 \\
\end{tabular}
}
\caption{ Metrics for medical parameter prediction from a single frame. Note: the video-based rPPG baseline results are reproduced as-reported from~\cite{egorov2025gaze} (train/test split unknown).}
\label{tab:results_medical}
\end{table}

\section{Discussion}
Our evaluation demonstrates that \modelname model surpasses conventional baselines in estimating basic biometric parameters, indicating its potential sustainability for medical-related applications.
While this potential requires validation through further experimentation, we argue that the proposed model can serve as new baseline for relevant studies, either as a frozen feature extractor or as a foundation for finetuning.
Consequently, the performance uplift cannot be attributed to the training dataset scale.

The core distinction between approach and it's closest analog, FaRL, lies in the training methodology.
FaRL employs a two-stage process: initial contrastive image-text alignment (CLIP-like training) followed by a masked image reconstruction task.
While this second stage shares conceptual ground with JEPA ideas, FaRL implements it through a separate autoencoder, calculating loss in that dedicated latent space rather than within the encoder's representation.
We contend that for medical applications, where predictive power is paramount, the FaRL framework has an inherent limitation.
Its initial contrastive stage is constrained by the quality and specificity of its text captions.
While subtle physiological cues can be described in principle, the captions available for large-scale web-scraped datasets like LAION almost never contain this level of medical detail.
Consequently, the model is optimized to align images with generic descriptions, which can dilute its focus on subtle, clinically relevant features. A pure self-supervised objective bypasses this constraint by learning directly from the visual data without relying on this intermediate, often irrelevant, textual signal.

Although we incorporate additional data sources, the final dataset for \modelname remains substantially smaller than the baseline's ones --- several time smaller than FaRL's FaceCaption-15M and about two orders of magnitude smaller than LAION-Faces-50M used for Franca.
Since all training sets derive from LAION~5B~\cite{relaion5b}, the efficiency gains must stem from our architectural modifications rather than data superiority.





\appendix
\section{Ablation studies}

For ablations studies, we selected BMI prediction as the most indicative task and used the small variant of \modelname model to reduce computational cost.
This work introduces several modifications to the classical JEPA pipeline: an axial stripe masking strategy,
a per-patch loss weighting scheme, and a probabilistic assignment of CLS token.
We ablate the impact of each component, with results summarized in the upper section of Table~\ref{tab:ablations}. The lower section of the table presents results for different values of \(P_{c\in\mathcal{S}}\) probability.

\begin{table}[htb]
\centering
\setlength{\colw}{1.4cm}
\begin{tabular}{c|c|c|c|c|c|c|c}
\hline
\multirow{2}{*}[-6pt]{Model} & \multirow{2}{*}[-6pt]{Size} & \multirow{2}{*}[-6pt]{Features} & \multirow{2}{*}[-6pt]{Masking} & \multirow{2}{*}[-6pt]{\(P_{c\in\mathcal{S}}\)} & \multirow{2}{*}[-2pt]{\shortstack{Weighted\\loss}} & \multicolumn{2}{c}{BMI} \\
\cline{7-8}
& & & & & & \makebox[\colw][c]{\(r^2\)} & \makebox[\colw][c]{MAE} \\
\hline
MeFEm & Small & Patches & Stripes \(2/{14}\)& \NA & Circular & 0.291 & 4.206 \\  
MeFEm & Small & Patches & Stripes \(3/{14}\) & \NA & Circular & \textbf{0.447} & 3.666 \\  
MeFEm & Small & Patches & Stripes \(4/{14}\)& \NA & Circular & 0.368 & 4.001 \\  
MeFEm & Small & Patches & Quadrants & \NA & Circular & 0.305 & 3.698 \\  
MeFEm & Small & Patches & Multiblcock & \NA & Circular & 0.371 & 3.961 \\  
MeFEm & Small & Patches & Stripes \(3/{14}\)& 0.5 & Uniform & 0.321 & 4.070 \\
MeFEm & Small & CLS & Stripes \(3/{14}\)& 0.5 & Uniform & 0.169 & 4.693 \\
MeFEm & Small & Patches + CLS & Stripes \(3/{14}\)& 0.5 & Uniform & 0.330 & 4.066 \\
\hline
MeFEm & Small & Patches & Stripes \(3/{14}\)& 0.0 & Circular & 0.388 & 3.752 \\
MeFEm & Small & CLS & Stripes \(3/{14}\)& 0.0 & Circular & 0.353 & 4.084 \\
MeFEm & Small & Patches + CLS & Stripes \(3/{14}\)& 0.0 & Circular & 0.378 & 3.697 \\
MeFEm & Small & Patches & Stripes \(3/{14}\)& 0.5 & Circular & 0.398 & 3.631 \\
MeFEm & Small & CLS & Stripes \(3/{14}\)& 0.5 & Circular & 0.430 & 3.781 \\
MeFEm & Small & Patches + CLS & Stripes \(3/{14}\)& 0.5 & Circular & 0.437 & \textbf{3.625} \\
MeFEm & Small & Patches & Stripes \(3/{14}\)& 1.0 & Circular & \textbf{0.444} & 3.631 \\
MeFEm & Small & CLS & Stripes \(3/{14}\)& 1.0 & Circular & 0.389 & 3.939 \\
MeFEm & Small & Patches + CLS & Stripes \(3/{14}\)& 1.0 & Circular & \textbf{0.446} & \textbf{3.627} \\
\end{tabular}
\caption{ Ablation results. Note: models in top/bottom sections were trained for 33/66 epochs respectively.}
\label{tab:ablations}
\end{table}

The ablation results indicate that the proposed axial stripe masking strategy is superior to both multiblock and quadrant masking alternatives.
Furthermore, the chosen strip width of 3 patches represents an optimum value, as performance degrades significantly with either decrease or increase from this value.
As expected, disabling the circular loss weighting also have negative impact, with CLS-only downstream metrics exhibiting a severe degradation.
This collapse in performance occurs because the CLS token is now forced to encode highly variable background information, a task that offers no theoretical or practical advantage for the target objective.

Considering probability \(P_{c\in\mathcal{S}}\) values, apart from balanced default value of \(0.5\) we additional considered two extremes: assigning CLS token exclusively to source set or using it only as prediction target.
Although models trained with \(P_{c\in\mathcal{S}}=1\) yield the highest raw metrics, they do so by sacrificing the predictive capability of the CLS token. 
In contrast, the value of \(P_{c\in\mathcal{S}}=0.5\) produce a more balanced model, with performance for patches embeddings and the CLS token approximately on par, while the case of \(P_{c\in\mathcal{S}}=0\) produces subpar results.
We see value in the balanced approach, as it allow for reduction in both downstream model size and training time while providing competitive results.
It may have additional merits, which which we will explore in the discussion.

\bibliography{bibliography}
\bibliographystyle{unsrt}

\end{document}